\documentclass[final]{cvpr}

\usepackage{times}
\usepackage{epsfig}
\usepackage{graphicx}
\usepackage{amsmath}
\usepackage{amssymb}
\usepackage{booktabs}
\usepackage{multirow}

\usepackage[pagebackref=true,breaklinks=true,colorlinks,bookmarks=false]{hyperref}

\def\cvprPaperID{10099} 
\def\confYear{CVPR 2021}
\newcommand{\absorption}{\sigma}
\newcommand{\absorptionavg}{\bar{\sigma}}

\newcommand{\radiance}{\mathbf{c}}
\newcommand{\radianceavg}{\bar{\mathbf{c}}}

\newcommand{\ray}{\mathbf{r}}
\newcommand{\pos}{\mathbf{x}}
\newcommand{\dir}{\mathbf{d}}
\newcommand{\rayorigin}{\mathbf{o}}
\newcommand{\rayparam}{t}
\newcommand{\rayparamnear}{t_n}
\newcommand{\rayparamfar}{t_f}
\newcommand{\transmittance}{T}
\newcommand{\transmittanceavg}{\bar{T}}

\newcommand{\render}{\mathbf{C}}
\newcommand{\renderapprox}{\mathbf{\tilde{C}}}

\newcommand{\length}{\delta}
\newcommand{\sampler}{\mathcal{S}}

\newcommand{\highlight}[1]{\textcolor{black}{#1}}

\newcommand{\imp}{\Phi}
\newcommand{\dimp}{\Psi}
\newcommand{\coords}{\bold{x}}
\newcommand{\coord}{x}
\newcommand{\param}{\theta}

\newcommand{\lin}[1]{\bold{W}_{#1}}

\newcommand{\nl}{{\textsc{nl}}}

\newcommand{\deriv}{\partial}
\newcommand{\dd}{\,\mathrm{d}}

\begin{document}

\title{AutoInt: Automatic Integration for Fast Neural Volume Rendering}

\author{David B. Lindell\thanks{Equal contribution. \newline\url{http://www.computationalimaging.org/publications/automatic-integration/}}
\qquad
Julien N. P. Martel\footnotemark[1] 
\qquad
Gordon Wetzstein\\[0.5em]
Stanford University\\
{\tt\small\{lindell, jnmartel, gordon.wetzstein\}@stanford.edu}
}

\maketitle

\begin{abstract}
Numerical integration is a foundational technique in scientific computing and is at the core of many computer vision applications. Among these applications, neural volume rendering has recently been proposed as a new paradigm for view synthesis, achieving photorealistic image quality. However, a fundamental obstacle to making these methods practical is the extreme computational and memory requirements caused by the required volume integrations along the rendered rays during training and inference. Millions of rays, each requiring hundreds of forward passes through a neural network are needed to approximate those integrations with Monte Carlo sampling.
Here, we propose automatic integration, a new framework for learning efficient, closed-form solutions to integrals using \highlight{coordinate-based neural networks.}
For training, we instantiate the computational graph corresponding to the derivative of the \highlight{network.}
The graph is fitted to the signal to integrate. After optimization, we reassemble the graph to obtain a network that represents the antiderivative. By the fundamental theorem of calculus, this enables the calculation of any definite integral in two evaluations of the network. 
\highlight{Applying this approach to neural rendering, we improve a tradeoff between rendering speed and image quality: improving render times by greater than 10$\times$ with a tradeoff of slightly reduced image quality.}
\end{abstract}


\vspace{-1em}
\section{Introduction}
\label{sec:introduction}
Image-based rendering and novel view synthesis are fundamental problems in computer vision and graphics (e.g.,~\cite{Carranza:2003,Szeliski:book}). The ability to interpolate and extrapolate a sparse set of images depicting a 3D scene has broad applications in entertainment, virtual and augmented reality, and many other applications. Emerging neural rendering techniques have recently enabled photorealistic image quality for these tasks (see Sec.~\ref{sec:related}).

Although state-of-the-art neural volume rendering techniques offer unprecedented image quality, they are also extremely slow and memory inefficient~\cite{mildenhall2020nerf}. This is a fundamental obstacle to making these methods practical. The primary computational bottleneck for neural volume rendering is the evaluation of integrals along the rendered rays during training and inference required by the volume rendering equation~\cite{max1995optical}.
Approximate integration using Monte Carlo sampling is typically used for this purpose, requiring hundreds of forward passes through the \highlight{neural network} representing the volume for each of the millions of rays that need to be rendered for a single frame.
Here, we develop a general and efficient framework for approximate integration. 
\highlight{Applied to the specific problem of neural volume rendering, our framework improves a tradeoff between rendering speed and image quality, allowing a greater than 10$\times$ speedup in the rendering process, though with some reduction in image quality.}

\begin{figure}[t!]
	\includegraphics[width=\columnwidth]{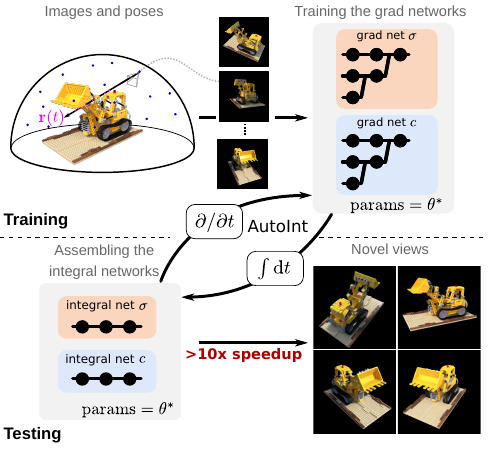}
	\caption{Automatic integration for neural volume rendering. During training, a grad network is optimized to represent multi-view images. At test time, we instantiate a corresponding integral network to rapidly evaluate per-ray integrals through the volume.}
	\label{fig:teaser}
    \vspace{-1em}
\end{figure}

Our integration framework builds on previous work demonstrating that \highlight{coordinate-based networks (sometimes also referred to as implicit neural representations)} can represent signals (e.g., images, audio waveforms, or 3D shapes) and their derivatives.
That is, taking the derivative of the \highlight{coordinate-based} network accurately models the derivative of the original signal.
This property has recently been shown for \highlight{coordinate-based networks} with periodic activation functions~\cite{sitzmann2020siren}, but we show that it also extends to a family of networks with different nonlinear activation functions (Sec~\ref{sec:ct} and supplemental).

We observe that taking the derivative of a \highlight{coordinate-based} network results in a new computational graph, a ``grad network'', which shares the parameters of the original network. 
Now, consider that we use as our network a multilayer perceptron (MLP).
Taking its derivative results in a grad network which can be trained on a signal that we wish to integrate.
By reassembling the grad network parameters back into the original MLP, we construct a neural network that represents the antiderivative of the signal to integrate.

This procedure results in a closed-form solution for the antiderivative, which, by the fundamental theorem of calculus, enables the calculation of any definite integral in two evaluations of the MLP.
Inspired by techniques for automatic differentiation (AutoDiff), we call this procedure \emph{automatic integration} or AutoInt.
Although the mechanisms of AutoInt and AutoDiff are very different, both approaches enable the calculation of integrals or derivatives in an automated manner that does not rely on traditional numerical techniques, such as sampling or finite differences.  

The primary benefit of AutoInt is that it allows evaluating arbitrary definite integrals quickly by querying the network representing the antiderivative. This concept could have important applications across science and engineering; here, we focus on the specific application of neural volume rendering. \highlight{For this application, efficiently evaluating integrals amounts to accelerating rendering (i.e., inference) times}, which is crucial for making these techniques more competitive with traditional real-time graphics pipelines. However, our framework still requires a slow training process to optimize a network for a given set of posed 2D images.  

Specifically, our contributions include the following.
\begin{itemize}
    \itemsep0em 
	\item We introduce a framework for automatic integration that learns closed-form integral solutions. To this end, we explore new network architectures and training strategies.
	\item Using automatic integration, we propose a new model and parameterization for neural volume rendering that is efficient in computation and memory.
    \highlight{\item We improve a tradeoff between neural rendering speed and image quality, demonstrating rendering rates that are an order of magnitude faster than previous implementations~\cite{mildenhall2020nerf}, though with a slight reduction in image quality.}
\end{itemize}

\section{Related Work}
\label{sec:related}
\paragraph{Neural Rendering.}

Over the last few years, end-to-end differentiable computer vision pipelines have emerged as a powerful paradigm wherein a differentiable or neural scene representation is optimized via differentiable rendering with posed 2D images (see e.g.,~\cite{tewari2020state} for a survey).
Neural scene representations often use an explicit 3D proxy geometry, such as multi-plane~\cite{flynn2019deepview,Mildenhall:2019,Zhou:2018} or multi-sphere~\cite{Attal:2020:ECCV,Broxton:2020} images or a voxel grid of features~\cite{Lombardi:2019,sitzmann2019deepvoxels}.
Explicit neural scene representations can be rendered quickly, but they are fundamentally limited by the large amount of memory they consume and thus may not scale well.

As an alternative, \highlight{coordinate-based networks, or implicit neural representations,} have been proposed as a continuous and memory-efficient approach.
Here, the scene is parameterized using neural networks, and 3D awareness is often enforced through inductive biases.
The ability to represent details in a scene is limited by the capacity of the network architecture rather than the resolution of a voxel grid, for example.
Such representations have been explored for modeling shape parts~\cite{genova2019deep,genova2019learning}, objects~\cite{atzmon2019sal,chabra2020deep,davies2020overfit,gropp2020implicit,kellnhofer2021neural,kohli2020inferring,liu2020dist,mescheder2019occupancy,michalkiewicz2019implicit,Niemeyer2020CVPR,Oechsle2019ICCV,park2019deepsdf,saito2019pifu,sitzmann2019srns,yariv2020multiview}, or scenes~\cite{eslami2018neural,jiang2020local,liu2020neural,mildenhall2020nerf,peng2020convolutional,sitzmann2020siren}.
\highlight{Coordinate-based networks} have also been explored in the context of generative frameworks~\cite{chan2020pi,chen2019learning,henzler2019platonicgan,hologan,nguyenphuoc2020blockgan,graf}.

The method closest to our application is neural radiance fields (NeRF)~\cite{mildenhall2020nerf}.
\highlight{NeRF is a neural rendering framework that combines a volume represented by a coordinate-based network with a neural volume renderer to achieve state-of-the-art image quality for view synthesis tasks.}
Specifically, NeRF uses ReLU-based multilayer perceptrons (MLPs) with a positional encoding strategy to represent 3D scenes.
Rendering an image from such a representation is done by evaluating the volume rendering equation~\cite{max1995optical}, which requires integrating along rays passing through the neural volume parameterized by the MLP.
This integration is performed using Monte Carlo sampling, which requires hundreds of forward passes through the MLP for each ray.
However, this procedure is extremely slow, requiring days to train a representation of a single scene from multi-view images.
Rendering a frame from a pre-optimized representation requires tens of seconds to minutes. 

Here, we leverage automatic integration, or AutoInt, to significantly speed up the evaluation of integrals along rays.
AutoInt reduces the number of network queries required to evaluate integrals (e.g., using Monte Carlo sampling) from hundreds to just two, \highlight{greatly speeding up inference for neural rendering}.

\begin{figure}[h!]
	\includegraphics[width=\columnwidth]{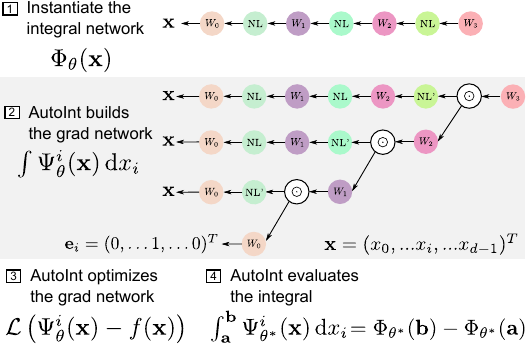}%
	\caption{AutoInt pipeline. After (1) defining an integral network architecture, (2) AutoInt builds the corresponding grad network, which is (3) optimized to represent a function. (4) Definite integrals can then be computed by evaluating the integral network, which shares parameters with its grad network.}%
	\label{fig:autoint}
    \vspace{-1.5em}
\end{figure}

\vspace{-1em}
\paragraph{Integration Techniques.}

In general, integration is much more challenging than differentiation. Whereas automatic differentiation primarily builds on the chain rule, there are many different strategies for integration, including variable substitution, integration by parts, partial fractions, etc. Heuristics can be used to choose one or a combination of these strategies for any specific problem. Closed-form solutions to finding general antiderivatives exist only for a relatively small class of functions and, when possible, involve a rather complex algorithm, such as the Risch or Risch-Norman algorithm~\cite{norman1977implementing,risch1969problem,risch1970solution}. Perhaps the most common approach to computing integrals in practice is numerical integration, for example using Riemann sums, quadratures, or Monte-Carlo methods~\cite{davis2007methods}. In these numerical methods, the number of samples trades off accuracy for runtime. 

Since neural networks are universal function approximators, and are themselves functions, they can also be integrated analytically. 
Previous work has explored theory and connections between shallow neural networks and integral formulations for function approximation~\cite{dereventsov2019neural,kainen2000integral}.
\highlight{Other work has derived closed-form solutions for integrals of simple single-layer or two-layer neural networks~\cite{teichert2019machine,turner2005introducing}.}
As we shall demonstrate, our work is not limited to a fixed number of layers or a specific architecture.
Instead, we directly train a grad network architecture for which the integral network is known by construction.

\section{\highlight{AutoInt for Neural Integration}}
\label{sec:integration}
In this section, we introduce a fundamentally new approach to compute and evaluate antiderivatives and definite integrals of \highlight{coordinate-based neural networks.}
\begin{figure*}[ht]
    \centering
	\includegraphics[width=\textwidth]{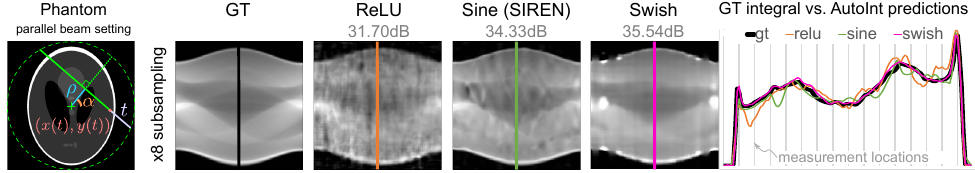}%
    \caption{AutoInt for computed tomography. Left: illustration of the parameterization. Center: sinograms computed for integral networks using different activation functions. In all cases, the ground truth (GT) sinogram is subsampled 8$\times$ and the optimized integral network is sampled to interpolate missing measurements. The Swish activation performs best  \highlight{in terms of peak signal-to-noise ratio (PSNR)}. Right: a 1D scanline of the sinogram shows that Swish interpolates missing data best while sine activations~\cite{sitzmann2020siren} tend to overfit the measurements.}%
	\label{fig:tomography}
	\vspace{-0.25cm}
\end{figure*}
\subsection{Principles}
We consider a \highlight{coordinate-based network}, i.e., a neural network with parameters $\theta$ mapping low-dimensional input coordinates to a low-dimensional output
$\imp_\param:\mathbb{R}^{d_{\mathrm{in}}}\mapsto\mathbb{R}^{d_{\mathrm{out}}}$.
We assume this \highlight{network} admits a (sub-)gradient with respect to its input $\coords\in\mathbb{R}^{d_{\mathrm{in}}}$, and we denote by $\dimp^i_\param=\deriv\imp_{\param}/\deriv{\coord_i}$ its derivative with respect to the coordinate $x_i$. 
Then, by the fundamental theorem of calculus we have that\vspace{-0.5em}
\begin{equation}
	\imp_\param(\coords)=\int \frac{\deriv{\imp_{\param}}}{\deriv{\coord_i}} (\coords)\dd{\coord_i} = \int \dimp^i_\param(\coords) \dd{\coord_i}.
\label{eqn:ftc}
\end{equation}
This equation relates the \highlight{coordinate-based network} $\imp_\param$ to its partial derivative $\dimp_\param^i$ and, hence, $\imp_\param$ is an antiderivative of $\dimp^i_\param$.

A key idea is that the partial derivative $\dimp^i_\param$ is itself \highlight{a coordinate-based network}, mapping the same low-dimensional input coordinates $\coords\in\mathbb{R}^{d_{\mathrm{in}}}$ to the same low-dimensional output space ${R}^{d_{\mathrm{out}}}$.
In other words, $\dimp^i_\param$ is a different neural network that shares its parameters $\theta$ with $\imp_\param$ while also satisfying Equation~\ref{eqn:ftc}. 
Now, rather than optimizing the \highlight{coordinate-based network} $\imp_\param$, we optimize $\dimp^i_\param$ to represent a target signal, and we reassemble the optimized parameters (i.e., weights and biases) $\param$ to form $\imp_\param$.

As a result, $\imp_\param$ is a network that represents an antiderivative of $\dimp^i_\param$. 
We call this procedure of training $\dimp^i_\param$ and reassembling $\param$ to construct the antiderivative \emph{automatic integration}. 
How to reassemble $\param$ depends on the network architecture used for $\imp_\param$, and is addressed in the next section.

\subsection{The Integral and Grad networks}
\highlight{Coordinate-based neural networks are usually formed from} multilayer perceptron (MLP), or fully connected, architectures:\vspace{-0.5em}
\begin{equation}
\imp_\theta(\coords)=\lin{n}(\phi_{n-1}\circ\phi_{n-2}\circ\dots\circ\phi_0)(\coords),
\end{equation}
with $\phi_k:\mathbb{R}^{M_k}\mapsto\mathbb{R}^{N_k}$ being the $k$-th layer of the neural network defined as $\phi_{k}(\bold{y})=\nl_k(\lin{k}\bold{y}+\bold{b}_k)$ 
using the parameters $\theta=\{ \lin{k}\in\mathbb{R}^{N_k\times M_k}, \bold{b}_k\in\mathbb{R}^{M_k}, \forall k \}$ and the nonlinearity $\nl$, which is a function applied point-wise to all the elements of a vector.

The computational graph of a 3-hidden-layer MLP representing $\imp_\theta$ is shown in Figure~\ref{fig:autoint}.
Operations are indicated as nodes and dependencies as directed edges. 
Here, the arrows of the directed edges point towards nodes that must be computed first. 

For this MLP, the form of the network $\dimp_\param^i=\deriv\imp_{\param}/\deriv{\coord_i}$ can be found using the \highlight{chain rule}\vspace{-0.5em}
\begin{align}
		\dimp_\param^i({\coords})= \hat{\phi}_{n-1}\circ(\phi_{n-2}\circ \dots \phi_0)(\bold{x})\odot \dots \nonumber \\ 
		\dots \odot\hat{\phi}_1\circ\phi_0(\bold{x}) \odot \lin{0}\,\bold{e}_i,
		\label{eq:grad_net}
\end{align}
where \highlight{$\odot$ indicates the Hadamard product,} $\hat{\phi}_k(\bold{y})=\lin{k}^T\nl_{k-1}'(\lin{k-1}\bold{y}+\bold{b}_{k-1})$ and $\bold{e}_i\in\mathbb{R}^{d_{\mathrm{in}}}$ is the unit vector that has $0$'s everywhere but at the $i$-th component.
The corresponding computational graph is shown in Figure~\ref{fig:autoint}.
As we noted, despite having a different architecture (and vastly different number of nodes) the two networks share the same parameters. 
We refer to the network associated with $\imp_\param$ as the \emph{integral network} and the neural network associated with $\dimp_\param^i$ as the \emph{grad network}.
Homologous nodes in their graphs are shown in the same color.
This color scheme explicitly shows how the grad network parameters are reassembled to create the integral network after training.

\subsection{Evaluating Antiderivatives \& Definite Integrals}

To compute the antiderivative and definite integrals of a function $f$ in the AutoInt framework, one first chooses the specifics of the MLP architecture (number of layers, number of features, type of nonlinearities) for an integral network $\Phi_\param$. 
The grad network $\dimp_\param^i$ is then instantiated from this integral network based on AutoDiff. 
In practice, we developed a custom AutoDiff framework that traces the integral network and explicitly instantiates the corresponding grad network while maintaining the shared parameters (additional details in the supplemental). 
Once instantiated, parameters of the grad network are optimized to fit a signal of interest using conventional AutoDiff and optimization tools~\cite{paszke2019pytorch}. Specifically we optimize a loss of the form
\begin{equation}
	\theta^* = \text{arg min}_\theta\: \mathcal{L}\left(\dimp^i_\param(\coords), f(\coords)\right).
\end{equation}
Here, $\mathcal{L}$ is a cost function that aims at penalizing discrepancies between the target signal $f(\coords)$ we wish to integrate and the \highlight{coordinate-based network} $\dimp_\param^i$.

Once trained, the grad network approximates the signal, that is 
$
	\dimp_{\param^*}^i \approx f(\coords), \:\forall\coords.
$
Therefore, the antiderivative of $f$ can be calculated as
\begin{equation}
	\int f(\coords) \dd\coord_i \approx \int \dimp_{\param^*}^i(\coords) \dd\coord_i = \Phi_{\theta^*}(\coords).
\end{equation}
This corresponds to evaluating the integral network at $\coords$ using weights $\param^*$. 
Furthermore, any definite integral of the signal $f$ can be calculated using \emph{only} two evaluations of $\imp_\param$, according to the Newton--Leibniz formula
\begin{equation}
	\int_{\bold{a}}^{\bold{b}} f(\coords) \dd{x_i} = \imp_\param(\bold{b})-\imp_\param(\bold{a}).
\end{equation}
We also note that AutoInt extends to integrating high-dimensional signals using a generalized fundamental theorem of calculus, which we describe in the supplemental.

\subsection{Example in Computed Tomography}
\label{sec:ct}
In tomography, integrals are at the core of the imaging model: measurements are line integrals of the absorption of a medium along rays that go through it. 
In particular, in a parallel beam setup, assuming a 2D medium of absorption $f(x,y)\in\mathbb{R}_+$, measurements can be modeled as
\begin{align}
	s(\rho,\alpha) = \int_{t_n}^{t_f} f\left(x(t),y(t)\right)\dd{t},
\end{align} 
and $(x,y)$ is on the ray $(\rho,\alpha)\in[-1,1]\times[0,\pi)$ by satisfying $x(t)\cos(\alpha) + y(t)\sin(\alpha) = \rho$ with $\alpha$ being the orientation of the ray and $\rho$ its eccentricity with respect to the origin as shown in Figure~\ref{fig:tomography}.
The measurement $s$ is called a sinogram, and this particular integral is referred to as the Radon transform of $f$~\cite{ct:book}.

The inverse problem of computed tomography involves recovering the absorption $f$ given a sinogram. Here, for illustrative purposes, we will look at a tomography problem in which a grad network is trained on a sparse set of measurements and the integral network is evaluated to produce unseen ones. \highlight{Sparse-view tomography is a standard reconstruction problem~\cite{gordon1970algebraic,pelt2018improving},} and this setup is analogous to the novel view synthesis problem we solve in Section~\ref{sec:rendering}. 

We consider a dataset of measurements $\mathcal{D}=\{(\rho_i,\alpha_i,s(\rho_i,\alpha_i)\}_{i<D}$ corresponding to $D$ sparsely sampled rays. We train a grad network using the AutoInt framework. For this purpose, we instantiate a grad network $\dimp_\param$ whose input is a tuple $(\rho,\alpha,t)$. It is trained to match the dataset of measurements 
\begin{equation}
	\theta^* = \text{arg min}_{\theta}\: \sum_{i<D} \Big\lVert \Big(\frac{1}{T}\sum_{t_j<T} \dimp^t_{\theta}(\rho_i,\alpha_i,t_j) \Big) - s(\rho_i,\alpha_i) \Big\rVert_2^2.
\end{equation}
Thus, at training time, the grad network is evaluated $T$ times in a Monte Carlo fashion with $t_j\sim\mathcal{U}([t_n,t_f])$. At inference, just two evaluations of $\imp_{\param^*}$ yield the integral
\begin{equation}
	s(\rho,\alpha)= \imp_{\param^*}(\rho,\alpha,t_f) - \imp_{\param^*}(\rho,\alpha,t_n).
\end{equation}
Results in Figure~\ref{fig:tomography} show that the two evaluations of the integral network $\imp_{\param^*}$ can faithfully reproduce supervised measurements and generalize to unseen data. 
Generalization, however depends on the type of nonlinearity used. 
We show that Swish~\cite{ramachandran2017searching} with normalized positional encoding (details in Sec.~\ref{sec:optimization}) generalizes well, and SIREN~\cite{sitzmann2020siren} fits the measurements better but fails to generalize to unseen views. 

Note that both the nonlinearity $\nl$ and its derivative $\nl'$ appear in the grad network architectures (Eq.~\eqref{eq:grad_net} and Figure~\ref{fig:autoint}).
This implies that integral networks with ReLU nonlinearities have step functions appearing in the grad network, possibly making training $\dimp_{\theta}$ difficult because of nodes with derivatives that are zero almost everywhere. 
We explore several other nonlinearities here (with additional details in the supplemental), and show that Swish heuristically performs best in the grad networks used in our application. Yet, we believe the study of nonlinearities in grad networks to be an important  avenue for future work.

\section{Neural Volume Rendering}
\label{sec:rendering}
Combining volume rendering techniques with \highlight{coordinate-based networks} has proved to be a powerful technique for neural rendering and view synthesis~\cite{mildenhall2020nerf}.
Here, we briefly overview volume rendering and describe an approximate volume rendering model that enables using AutoInt for efficient rendering.

\subsection{Volume Rendering}
Classical volume rendering techniques are derived from the radiative transfer equation~\cite{chandrasekhar2013radiative} with an assumption of minimal scattering in an absorptive and emissive medium~\cite{drebin1988volume,max1995optical}.
We adopt a rendering model based on tracing rays through the volume~\cite{kajiya1984ray, mildenhall2020nerf}, where the emission and absorption along camera rays produce color values that are assigned to rendered pixels. 

The volume itself is represented as a high-dimensional function parameterized by position, $\pos\in\mathbb{R}^3$, and viewing direction $\dir$.
We also define the camera rays that traverse the volume from an origin point $\rayorigin$ to a ray position $\ray(\rayparam)=\rayorigin + t\,\dir$.
At each position in the volume, an absorption coefficient, \highlight{$\absorption\in\mathbb{R}_+$}, gives the probability per differential unit length that a ray is absorbed (i.e., terminates) upon interaction with an infinitesimal particle. 
Finally, an emissive radiance field \highlight{$\radiance=(r,g,b)\in[0, 1]^3$}, describes the color of emitted light at each point in space in all directions. 

Rendering from the volume requires integrating the emissive radiance along the ray while also accounting for absorption. 
The transmittance $\transmittance$ describes the net reduction from absorption from the ray origin to the ray position $\ray(\rayparam)$, and is given as \vspace{-1em}
\begin{equation}
    \transmittance(\rayparam) = \exp\left(-\int_{\rayparamnear}^\rayparam \absorption\big(\ray(s)\big) \dd{s}\right),
\end{equation}
where $\rayparamnear$ indicates a near bound along the ray.
With this expression, we can define the volume rendering equation \highlight{(VRE)}, which describes the color $\render$ of a rendered camera ray.\vspace{-1em}
\begin{align}
    \render(\ray) = \int_{\rayparamnear}^{\rayparamfar} \transmittance(\rayparam) \, \absorption(\ray(\rayparam)) \, \radiance(\ray(\rayparam), \mathbf{\dir}) \dd{t}.
\end{align}
Conventionally, the \highlight{VRE} is computed numerically by Riemann sums, quadratures, or Monte-Carlo methods~\cite{davis2007methods}, whose accuracy thus largely depends on the number of samples taken along the ray. 

\subsection{Approximate Volume Rendering for Automatic Integration}
Automatic integration allows us to efficiently evaluate definite integrals using a closed-form solution for the antiderivative.
However, the \highlight{VRE} cannot be directly evaluated with AutoInt because it consists of multiple nested integrations: the integration of radiance along the ray weighted by integrals of cumulative transmittance.
We therefore choose to approximate this integral in piecewise sections that can each be efficiently evaluated using AutoInt.
For $N$ piecewise sections along a ray, we give the approximate \highlight{VRE} and transmittance as\vspace{-0.5em}
\begin{align}
    \renderapprox(\ray) =\sum_{i=1}^N \absorptionavg_i \, \radianceavg_i \transmittanceavg_i\,\delta_i, \quad{}
    \transmittanceavg_i = \exp\left( -\sum_{j=1}^{i-1} \bar{\sigma}_j \delta_j\right), \label{eqn:approx_vre}
\end{align}
where \vspace{-0.5em}
\begin{align}
    \absorptionavg_i = \length_i^{-1}\int_{t_{i-1}}^{t_i} \absorption(\rayparam) \dd{t}
    \quad{}\text{and}\quad{}
    \radianceavg_i = \length_i^{-1} \int_{t_{i-1}}^{t_i} \radiance(\rayparam) \dd{t}, \nonumber
    \vspace{-1em}
\end{align}
and $\length_i = t_i - t_{i-1}$ is the length of each piecewise interval along the ray.
\highlight{Equation~\ref{eqn:approx_vre} can also be viewed as a repeated alpha compositing operation with alpha values of $\absorptionavg_i \delta_i$.}
After some simplification and substitution into Equation~\ref{eqn:approx_vre} (see supplemental), we have the following expression for the piecewise \highlight{VRE}:
\begin{align}
    \renderapprox(\ray) = \sum\limits_{i=1}^N \underbrace{\length_i^{-1}\int_{t_{i-1}}^{t_i} \absorption(\rayparam) \, \mathrm{d}t}_{\absorptionavg_i} \cdot\underbrace{\int_{t_{i-1}}^{t_i} \radiance(\rayparam) \dd{t}}_{\radianceavg_i\length_i} \label{eqn:vre_piecewise} \\
    \cdot\underbrace{\prod\limits_{j=1}^{i-1} \exp\left(-\int_{t_{j-1}}^{t_j} \absorption(s) \, \dd{s}\right)}_{\transmittanceavg_i}.\nonumber
\vspace{-0.5em}
\end{align}
While this piecewise expression is only an approximation to the full \highlight{VRE}, it enables us to use AutoInt to efficiently evaluate each piecewise integral over absorption and radiance. 
In practice, there is a tradeoff between improved computational efficiency and degraded accuracy of the approximation as the value of $N$ decreases.
We evaluate this tradeoff in the context of volume rendering and learned novel view synthesis in Sec.~\ref{sec:results}. 

%
\begin{figure*}[t!]
	\includegraphics[width=\textwidth]{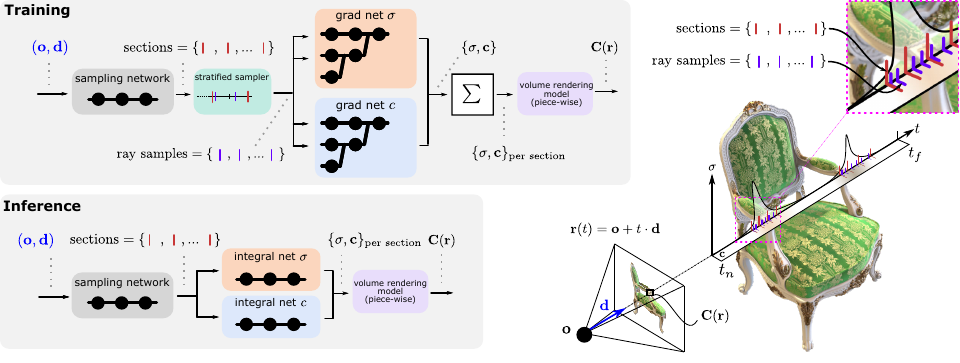}%
	\caption{Volume rendering pipeline. During training, the grad networks representing volume density $\absorption$ and color $\radiance$ are optimized for a given set of multi-view images (top left). For inference, the grad networks' parameters are reassembled to form the integral networks, which represent antiderivatives that can be efficiently evaluated to calculate ray integrals through the volume (bottom left). A sampling network predicts the locations of piecewise sections used for evaluating the definite integrals (right).}%
	\label{fig:volumetric_rendering}
\vspace{-1em}
\end{figure*}
%

\vspace{-0.5em}
\section{Optimization Framework}
\label{sec:optimization}
We evaluate the piecewise \highlight{VRE} introduced in the previous section using an optimization framework overviewed in Figure~\ref{fig:volumetric_rendering}.  
At the core of the framework are two MLPs that are used to compute integrals over values of $\absorption$ and $\radiance$ as we detail in the following.

\vspace{-1em}
\paragraph{Network Parameterization.}

Rendering an image from the high-dimensional volume represented by the MLP requires evaluating integrals along each ray  $\ray(\rayparam) $ in the direction of $\rayparam$. 
Thus, the grad network should represent $\partial\imp_\param/\partial \rayparam$, the partial derivative of the integral network with respect to the ray parameter.  
In practice, the networks take as input the values that define each ray: $\rayorigin$, $\rayparam$, and $\dir$.
Then, positions along the ray are calculated as $\pos = \rayorigin + \rayparam\,\dir$ and passed to the initial layers of the networks together with $\dir$.
With this dependency on $\rayparam$, we use our custom AutoDiff implementation to trace computation through the integral network, define the computational graph that computes the partial derivative with respect to $\rayparam$, and instantiate the grad network.

\vspace{-1.0em}
\paragraph{Grad Network Positional Encoding.}

As demonstrated by Mildenhall et al.~\cite{mildenhall2020nerf}, a positional encoding on the input coordinates to the network can significantly improve the ability of a network to render fine details. 
We adopt a similar scheme, where each input coordinate is mapped into a higher dimensional space as using a function $\gamma(p): \mathbb{R}\mapsto\mathbb{R}^{2L}$ defined as 
\begin{equation}
    \resizebox{\columnwidth}{!}
    {%
        $\gamma(p) = \left(\sin(\omega_0 p), \cos(\omega_0 p), \cdots, \sin(\omega_{L-1} p), \cos(\omega_{L-1} p)\right)$,%
    }
\end{equation}
where $\omega_i=2^i\pi$ and $L$ controls the number of frequencies used to encode each input.
We find that using this scheme directly in the grad network produces poor results because it introduces an exponentially increasing amplitude scaling into the coordinate encoding. 
This can easily be seen by calculating the derivative $\partial\gamma/\partial p=\left(\cdots \omega_i\cos(\omega_i p), -\omega_i\sin(\omega_i p) \cdots\right)$.
Instead, we use a normalized version of the positional encoding for the integral network, which improves performance when training the grad network:
\begin{equation}
        \bar{\gamma}(p) = \left(\cdots, \omega_i^{-1}\sin(\omega_i p), \omega_i^{-1}\cos(\omega_i p), \cdots \right).
\end{equation}

\vspace{-1.5em}
\paragraph{Predictive Sampling.}
While AutoInt is used at inference time, at training time, the grad network is optimized by evaluating the piecewise integrals of Equation~\ref{eqn:vre_piecewise} using a quadrature rule discussed by Max~\cite{max1995optical}:
\begin{align}
    \renderapprox(\ray) = \sum\limits_{i=1}^N \transmittanceavg_i \left(1 - \exp(-\absorptionavg_i\length_i) \right)\radianceavg_i\,.
\end{align}
We use Monte Carlo sampling to evaluate the integrals $\absorptionavg_i$ and $\radianceavg_i$ by querying the networks at many positions within each interval $\length_i$.

However, some intervals $\length_i$ along the ray contribute more to a rendered pixel than others.
Thus, assuming we use the same number of samples per interval, we can improve sample efficiency by strategically adjusting the length of these intervals to place more samples in positions with large variations in $\absorption$ and $\radiance$. 
This idea is similar in spirit to accelerated volume rendering techniques using hierarchical sampling or adaptive ray termination~\cite{levoy1990efficient}. 

To this end, we jointly train a small sampling network (illustrated in Figure~\ref{fig:volumetric_rendering}), which is implemented as an MLP $\sampler(\rayorigin, \dir)$ that predicts interval lengths $\boldsymbol{\length}\in\mathbb{R}^N$.
Then, we calculate stratified samples along the ray by subdividing each interval $\length_i$ into $M$ bins and calculating samples $t_{i,j}, \, j=1, \ldots, M$ as $t_{i,j} \sim \mathcal{U}\left(t_{i-1} + \frac{j-1}{M}\length_i, t_{i-1} + \frac{j}{M}\length_i \right)$.

\vspace{-1em}
\paragraph{Fast Grad Network Evaluation.}
AutoInt can be implemented directly in popular optimization frameworks (e.g., PyTorch~\cite{paszke2019pytorch}, Tensorflow~\cite{tensorflow2015-whitepaper}); however, training the grad network is generally computationally slow and memory inefficient.
These inefficiencies stem from the two step procedure required to compute the grad network output at each training iteration: (1) a forward pass through the integral network is computed and then (2) AutoDiff calculates the derivative of the output with respect to the input variable of integration.
Instead, we implemented a custom AutoDiff framework on top of PyTorch that parses a given integral network and explicitly instantiates the grad network modules with weight sharing (see Figure~\ref{fig:autoint}).
Then, we evaluate and train the grad network directly, without the overhead of the additional per-iteration forward pass and derivative computation. 
Compared to the two-step procedure outlined above, our custom framework improves per-iteration training speed by a factor of 1.8  and reduces memory consumption by 15\% for the volume rendering application. 
\highlight{More details about our AutoInt implementation can be found in the supplemental, and our code is publicly available\footnote{\url{https://github.com/computational-imaging/automatic-integration}}.}

\begin{figure}
	\includegraphics[scale=1]{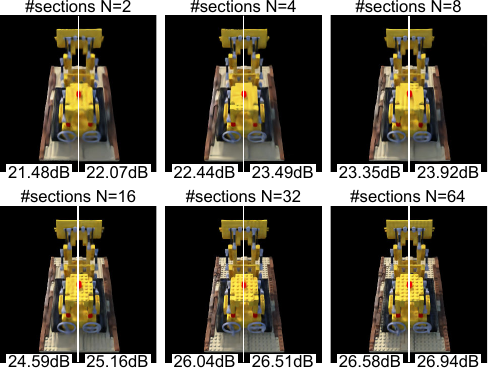}
    \caption{Ablation studies. A view of the \textit{Lego} scene is shown with a varying number of intervals ($N=\{2,4,8,16,32,64\}$) without (left half of the images) and with (right half) the sampling network. PSNR is computed on the 200 test set views.}
	\label{fig:ablation}
	\vspace{-1.5em}
\end{figure}
\vspace{-1em}
\paragraph{Implementation Details.}
In our framework, a volume representation is optimized separately for each rendered scene.
To optimize the grad networks, we require a collection of RGB images taken of the scene from varying camera positions, and we assume that the camera poses and intrinsic parameters are known.
At training time, we randomly sample images from the training dataset, and from each image we randomly sample a number of rays.
Then, we optimize the network to minimize the loss function 
\vspace{-0.5em}
\begin{equation}
    \mathcal{L} = \sum\limits_{\ray} \lVert \renderapprox(\ray) - \render(\ray)\rVert_2^2,
    \vspace{-0.5em}
\end{equation}
where $\render$ is the ground truth pixel value for the selected ray.

In our implementation, we train the networks using PyTorch and the Adam optimizer~\cite{kingma2014adam} with a learning rate of $5\times 10^{-4}$.
\highlight{The networks representing volume density and color each have 8 hidden layers with 256 hidden units,} we use a batch size of 4 with 1024 rays sampled from each image, and we decay the learning rate by a factor of 0.2 every $10^5$  iterations. 
\highlight{Training and inference are performed using NVIDIA V100 GPUs.}
For the sampling network, we evaluate using $M = 128/N$ samples within each piecewise interval for $N\in\{2, 4, 8, 16, 32, 64\}$ (see Figure~\ref{fig:ablation}) and find that using 8, 16, or 32 piecewise intervals produces acceptable results while achieving a significant computational acceleration with AutoInt.
Finally, for the positional encoding, we use $L=10$ and $L=4$ for $\pos$ and $\dir$, respectively.

\vspace{-0.5em}
\section{Results}
\label{sec:results}
\begin{figure*}[ht]
    \centering
	\includegraphics[scale=0.98]{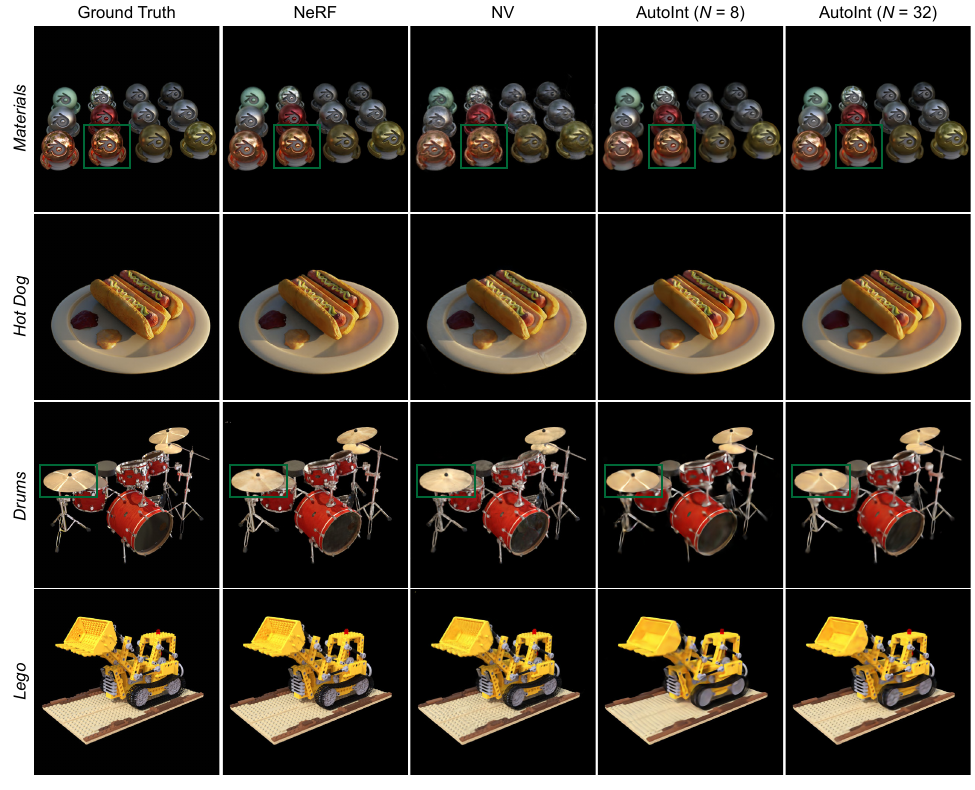}
    \caption{Qualitative results. We compare the performance of Neural Volumes~\cite{Lombardi:2019} and NeRF~\cite{mildenhall2020nerf} to AutoInt using $N=8$ and $N=32$ in our approximate volume rendering equation. AutoInt accurately captures view-dependent effects like specular reflections (green boxes) and reduces render times by greater than 10$\times$ relative to NeRF, though with some reduction in overall image quality.} 
    \label{fig:baselines}
    \vspace{-1.5em}
\end{figure*}

\begin{table}
\resizebox{\columnwidth}{!}{
\centering
	\begin{tabular}{r c c ccc}
		\toprule[0.8pt]
		& \multirow{2}{*}{NeRF} & \multirow{2}{*}{\shortstack{Neural\\Volumes}}  
		& \multicolumn{3}{c}{AutoInt ($N$=\#sections)}\\
		\cmidrule[0.2pt]{4-6}
		& 	   &   & $N=8$ & $N=16$ & $N=32$ \\
		\midrule[0.4pt]
PSNR (dB)	& 31.0 & 26.1 & 25.6 & 26.0 & 26.8  \\
Memory (GB)	& 15.6 & 10.4 & 15.5 & 15.0 & 15.5 \\
Runtime (s/frame) & 30 & 0.3 & 2.6 & 4.8 & 9.3 \\
		\bottomrule[0.8pt]
	\end{tabular}
}
    \vspace{0.1em}
	\caption{NeRF~\cite{mildenhall2020nerf} achieves the best image quality measured by average peak signal-to-noise ratio (PSNR). Neural Volumes~\cite{Lombardi:2019} is faster and slightly more memory efficient, but suffers from lower image quality. AutoInt allows us to approximate the NeRF solution with a tradeoff between image quality and runtime defined by the number of intervals used by our sampling network. Results are aggregated over the 8 Blender scenes of the NeRF dataset. }
	\label{tab:recap}
	\vspace{-0.25cm}
\end{table}
\begin{table}
\resizebox{\columnwidth}{!}{
\centering
	\begin{tabular}{cccccc}
		\toprule[0.8pt]
        VRE & Network Type & Samples/Forward Passes & PSNR (dB) $\uparrow$ & SSIM $\uparrow$ & LPIPS $\downarrow$\\
        \midrule[0.4pt]
        \multirow{3}{*}{\shortstack{Piecewise\\(approx.)}} & \multirow{3}{*}{\shortstack{Grad MLP\\(proposed)}} & 9, $N=8$ & 25.09 & 0.900 & 0.175 \\
        & & 17, $N=16$  & 25.48  & 0.905  & 0.171 \\
        & & 33, $N=32$  & 27.26 & 0.929 & 0.135 \\
        \midrule[0.4pt]
        \multirow{3}{*}{\shortstack{Piecewise\\(approx.)}} & \multirow{3}{*}{Standard MLP} & 128, $N=8$ & 29.21 & 0.952 & 0.052 \\
        & & 128, $N=16$  & 29.97 & 0.960 & 0.047 \\
        & & 128, $N=32$  & 29.68 & 0.959 & 0.049 \\
        \midrule[0.4pt]
        Full (exact) & Grad MLP & 128 & 27.95 & 0.936 & 0.082\\
        \midrule[0.4pt]
        \multicolumn{2}{c}{\multirow{3}{*}{NeRF}} & 128 & 30.68 & 0.968 & 0.045\\
                                                  & & 32 & 23.30 & 0.920 & 0.093\\
                                                  & & 8 & 14.62 & 0.761 & 0.258\\
		\bottomrule[0.8pt]
    \end{tabular}}
    \vspace{0.1em}
    \caption{\highlight{Comparison of performance on the \textit{Lego} scene for different network configurations. We report PSNR/SSIM~\cite{wang2004image} and LPIPS~\cite{zhang2018unreasonable}. AutoInt uses the piecewise VRE and a grad network (top rows), and the number of forward passes required at inference depends on the number of piecewise sections ($N$). We also evaluate using a fixed number of samples with the piecewise VRE and a standard MLP (i.e., Monte Carlo sampling, no grad network), as well as using the full VRE with a grad network. Finally we compare to NeRF~\cite{mildenhall2020nerf} using varying samples at inference, which reduces computational requirements.}}
	\label{tab:ablation}
    \vspace{-1.5em}
\end{table}
We evaluate AutoInt for volume rendering on a synthetic dataset of scenes with challenging geometries and reflectance properties. 
\highlight{We demonstrate that the approach allows an improved tradeoff between image quality and rendering speed for neural volume rendering. Rendering times are improved by greater than 10$\times$ compared to the state-of-the-art~\cite{mildenhall2020nerf}, though at slightly reduced image quality.}

Our training dataset consists of eight objects, each rendered from 100 different camera positions using the Blender Cycles engine~\cite{mildenhall2020nerf}.
For the test set, we evaluate on an additional 200 images.
We compare AutoInt to two other baselines: Neural Radiance Fields (NeRF)~\cite{mildenhall2020nerf} and Neural Volumes~\cite{Lombardi:2019}.
NeRF uses a similar architecture and Monte Carlo sampling with the full volume rendering model, rather than our piecewise approximation and AutoInt. 
Neural Volumes is a voxel-based method that encodes a deep voxel grid representation of a scene using a convolutional neural network.
Novel views are rendered by applying a learned warping operator to the voxel grid and sampling voxel values by marching rays from the camera position.

In Table~\ref{tab:recap} we report the peak signal-to-noise ratio (PSNR) averaged across all scenes and test images.
AutoInt outperforms Neural Volumes quantitatively, while achieving a greater than 10$\times$ improvement in render time relative to NeRF\highlight{, though with a tradeoff in image quality}.
\highlight{Increasing the number of piecewise sections in the approximate VRE improves render quality at the cost of computation.} 

\highlight{We evaluate the effect of the sampling network and the number of sections in the approximate VRE in Figure~\ref{fig:ablation} for the \emph{Lego} scene. Using the sampling network improves performance and sample efficiency by allocating more sections in regions with large variations in the volume density.}

\highlight{In Table~\ref{tab:ablation} we show the effect of the VRE approximation and grad network architecture on render quality of the \textit{Lego} scene. Using the full VRE achieves similar performance to the approximate, piecewise VRE with 32 sections. We attribute most of the difference in performance between our method and NeRF to the regularized, tree-like structure of the grad network, which is constrained by weight sharing between the branches (see Figure~\ref{fig:autoint}). While evaluating NeRF with fewer samples along each ray reduces computation, rendering quality degrades significantly compared to using AutoInt with the same number of samples.}

\highlight{We also show qualitative results in Figure~\ref{fig:baselines} for the \emph{Materials}, \emph{Hot Dog}, \emph{Drums}, and \emph{Lego} scenes.}
Again, the quality of the rendered images improves as the number of sections increases.
\highlight{In the \emph{Materials} scene (Figure~\ref{fig:baselines}), the proposed technique exhibits fewer artifacts compared to Neural Volumes.}
\highlight{AutoInt also shows improved modeling of view-dependent effects in the \emph{Drums} scene relative to Neural Volumes and NeRF (e.g., specular highlights on the symbols).}
\highlight{We show additional results on captured scenes from the Local Light Field Fusion and DeepVoxels datasets~\cite{Mildenhall:2019,sitzmann2019deepvoxels} in the supplemental}.

\vspace{-0.5em}
\section{Discussion}
\label{sec:discussion}
In this work, we introduce a new framework for numerical integration in the context of \highlight{coordinate-based neural networks}. 
Applied to neural volume rendering, AutoInt enables improvements to computational efficiency by learning closed-form solutions to integrals.
\highlight{Although these computational speedups currently come with a tradeoff to image quality, the method takes steps towards efficient learned integration using deep network architectures.} 

Our approach is analogous to conventional methods for fast evaluation of the \highlight{VRE}; for example, methods based on shear-warping~\cite{lacroute1994fast} and the Fourier projection-slice theorem~\cite{malzbender1993fourier,totsuka1993frequency}.
Similar to our method, these techniques use approximations (e.g., with sampling and interpolation) that trade off image quality with computationally efficient rendering.
Additionally, we believe our approach is compatible with recent work that aims to speed up volume rendering by pruning areas of the volume that do not contain the rendered object~\cite{liu2020neural}.  

A key idea of AutoInt is that an integral network can be automatically created after training a corresponding grad network.
We note that grad networks represent a fundamentally different and relatively unexplored type of network architecture compared to conventional fully-connected networks. 
While using grad networks with certain classes of non-linearities (e.g., Swish) can improve image quality, exploring improved training strategies and more expressive grad network architectures is an important and promising direction for future work. 
Finally, we believe that AutoInt will be of interest to a wide array of application areas beyond computer vision, especially for problems related to inverse rendering, sparse-view tomography, and compressive sensing.

\paragraph{Acknowledgments.} Julien N. P. Martel was supported by a Swiss National Foundation (SNF) Fellowship (P2EZP2 181817). Gordon Wetzstein was supported by an NSF CAREER Award (IIS 1553333) and a PECASE from the ARO.

{\small
\bibliographystyle{ieee_fullname}
\bibliography{references}
}

\end{document}



\title{Supplemental Material\\AutoInt: Automatic Integration for Fast Neural Volumetric Rendering}

\author{David B. Lindell\thanks{Equal contribution. \newline\url{http://www.computationalimaging.org/publications/automatic-integration/}}
\qquad
Julien N. P. Martel\footnotemark[1] 
\qquad
Gordon Wetzstein\\[0.5em]
Stanford University\\
{\tt\small\{lindell, jnmartel, gordon.wetzstein\}@stanford.edu}
}

\maketitle


\section{Multivariable Integration with AutoInt}
We consider an implicit neural representation realized by a neural network with parameters $\theta$.
The network maps low-dimensional input coordinates to a low-dimensional output
$\imp_\param:\mathbb{R}^{d_{\mathrm{in}}}\mapsto\mathbb{R}^{d_{\mathrm{out}}}$, and we assume that the network admits a (sub-)gradient with respect to its input $\coords\in\mathbb{R}^{d_{\mathrm{in}}}$.
We denote by $\dimp^i_\param=\deriv\imp_{\param}/\deriv{\coord_i}$ the derivative of the network output with respect to the input coordinate $x_i$, and, as described in the main text, we call $\dimp^i_\theta$ the grad network and $\imp_\theta$ the integral network.

By the fundamental theorem of calculus, the grad network and integral network are related as
\begin{equation}
	\imp_\param(\coords)=\int \frac{\deriv{\imp_{\param}}}{\deriv{\coord_i}} (\coords)\dd{\coord_i} = \int \dimp^i_\param(\coords) \dd{\coord_i}.
\label{eqn:supp_ftc}
\end{equation}
As a corollary we have that definite integrals can be computed by two evaluations of the integral network: 
\begin{equation}
	\int_{a_i}^{b_i} \dimp^i_\param(\coords) \dd{\coord_i} = \imp_\theta(\coords)\Big\vert_{x_i=b_i} - \imp_\theta(\coords)\Big\vert_{x_i=a_i}.
\label{eqn:supp_definite}
\end{equation}

Now, we will extend this result to multiple integrations.
First, we let $\dimp^{i,j}_\param=\deriv\dimp^{i}_{\param}/\deriv{\coord_j}$ be the partial derivative with respect to $x_j$ such that
\begin{equation}
	\dimp_\theta^{i} = \int \frac{\deriv{\dimp^{i}_{\param}}}{\deriv{\coord_j}} (\coords)\dd{\coord_j} = \int \dimp^{i,j}_\param(\coords) \dd{\coord_j}.
\label{eqn:supp_ftc2}
\end{equation}
Then we can express the double integral as
\begin{multline}
	\int_{a_i}^{b_i}\int_{a_j}^{b_j} \dimp^{i,j}_\param(\coords) \dd{\coord_j}\dd{\coord_i} \\
	= \int_{a_i}^{b_i} \dimp^{i}_\param(\coords)\Big\vert_{x_j=b_j} - \dimp^{i}_\param(\coords)\Big\vert_{x_j=a_j} \dd{\coord_i}\\
        = \left(\imp_\param(\coords)\Big\vert_{x_j=b_j} - \imp_\param(\coords)\Big\vert_{x_j=a_j} \right)\Big\rvert_{x_i=b_i}  \\
        - \left(\imp_\param(\coords)\Big\vert_{x_j=b_j} - \imp_\param(\coords)\Big\vert_{x_j=a_j} \right)\Big\rvert_{x_i=a_i}.
\label{eqn:supp_definite2}
\end{multline}
Equation~\ref{eqn:supp_definite2} can be further simplified with a slight abuse of notation by letting $\imp_\param(\coords)\Big\vert_{x_i=a_i, x_j=a_j} = \imp_\param(a_i, a_j)$, resulting in
\begin{align}
\int_{a_i}^{b_i}\int_{a_j}^{b_j} \dimp^{i,j}_\param(\coords) \dd{\coord_j}\dd{\coord_i} 
&= \imp_\param(b_i, b_j) - \imp_\param(b_i, a_j) \nonumber \\
&- \imp_\param(a_i, b_j) + \imp_\param(a_i, a_j).
\end{align}
Stated otherwise, a definite integral over two dimensions can be computed with four evaluations of the integral network at the given bounds.

This result can be extended to $n$ dimensions with the following formula~\cite{mutze2010}.
\begin{align}
&\int_{a_1}^{b_1}\cdots\int_{a_n}^{b_n} \dimp^{1,\ldots,n}_\param(\coords) \dd{\coord_n}\cdots\dd{\coord_1} \nonumber\\
&= \hspace{-1em}\sum\limits_{\epsilon_1, \ldots, \epsilon_n=0}^{1} (-1)^{\epsilon_1 + \cdots + \epsilon_n} \imp_\theta(\epsilon_1 a_1 + \bar{\epsilon}_1 b_1, \ldots, \epsilon_n a_n + \bar{\epsilon}_n b_n),
\label{eqn:ndim}
\end{align}
with $\bar{\epsilon}_i = 1 - \epsilon_i$. Thus for $n=3$, we would have 
\begin{align}
    &\quad\imp_\theta(b_i, b_j, b_k) - \imp_\theta(b_i, a_j, b_k) - \imp_\theta(a_i, b_j, b_k) \nonumber \\
    &+ \imp_\theta(a_i, a_j, b_k) - \imp_\theta(b_i, b_j, a_k) + \imp_\theta(b_i, a_j, a_k) \nonumber \\
    &+ \imp_\theta(a_i, b_j, a_k) - \imp_\theta(a_i, a_j, a_k).
\end{align}

Overall, using AutoInt for multivariable integration follows a similar procedure to evaluating a single integral as described in the main text. 
First, one constructs the grad network by taking partial derivatives of the integral network with respect to each of the variables of integration. 
Then, after training, the integral network is reassembled from the parameters $\param$ and evaluated at the bounds of the domain as described by Equation~\ref{eqn:ndim}.

\section{Deriving the VRE Approximation and Quadrature}
\subsection{Piecewise Constant VRE}
Our approximation of the volume rendering equation (VRE), can be viewed as a Riemann integral using $N$ piecewise constant sections:
\begin{align}
    \renderapprox(\ray) =\sum_{i=1}^N \absorptionavg_i \, \radianceavg_i \transmittanceavg_i\,\delta_i,\quad{}
     \label{eqn:approx_vre}
\end{align}
where $\length_i = t_i - t_{i-1}$ is the length of the section $i$ along the ray. The density $\bar{\sigma}_i$ and radiance $\bar{c}_i$ of each section are defined as:
\begin{align}
    \absorptionavg_i = \length_i^{-1}\int_{t_{i-1}}^{t_i} \absorption(\rayparam) \dd{t},
    \label{eqn:avg_density}
\end{align}
and
\begin{align}
    \radianceavg_i = \length_i^{-1} \int_{t_{i-1}}^{t_i} \radiance(\rayparam) \dd{t},
    \label{eqn:avg_rad}
\end{align}
and the transmittance as
\begin{align}
	\transmittanceavg_i = \exp\left( -\sum_{j=1}^{i-1} \bar{\sigma}_j \delta_j\right).
	\label{eqn:avg_trans}
\end{align}
After substituting the terms defined in Equations~\eqref{eqn:avg_density},\eqref{eqn:avg_rad}~and~\eqref{eqn:avg_trans} into Equation~\eqref{eqn:approx_vre} and simplifying, we obtain the following expression for the piecewise volume rendering equation:
\begin{align}
    \renderapprox(\ray) = \sum\limits_{i=1}^N \length_i^{-1}\int_{t_{i-1}}^{t_i} \absorption(\rayparam) \, \mathrm{d}t \cdot\int_{t_{i-1}}^{t_i} \radiance(\rayparam) \dd{t} \label{eqn:vre_piecewise} \\
\cdot\prod\limits_{j=1}^{i-1} \exp\left(-\int_{t_{j-1}}^{t_j} \absorption(s) \, \dd{s}\right).\nonumber
\end{align}
\subsection{Calculation via Quadrature}
To obtain the quadrature rule proposed by Max~\cite{max1995optical}, used in the NeRF model, we first note that in the limit of $\absorptionavg_i\length_i\rightarrow 0$ the following Taylor expansion holds
\begin{equation}
	\exp(\absorptionavg_i\length_i)\overset{\absorptionavg_i\length_i\rightarrow 0}{=} 1 + \absorptionavg_i\length_i + O(\absorptionavg_i\length_i).
\end{equation}
That is, using the definition of $\bar{\sigma}_i$ from Equation~\eqref{eqn:avg_density} we have
\begin{equation}
	1-\exp(-\absorptionavg_i\length_i) \approx \absorptionavg_i\length_i = \int_{t_{i-1}}^{t_i} \absorption(\rayparam) \dd{\rayparam},
\end{equation}
yielding the quadrature
\begin{equation}
	\renderapprox(\ray) = \sum\limits_{i=1}^N \transmittanceavg_i \left(1 - \exp(-\absorptionavg_i\length_i) \right)\radianceavg_i\,	
\end{equation}

\section{AutoInt Implementation}
\begin{figure*}[h!]
	\includegraphics[width=\textwidth]{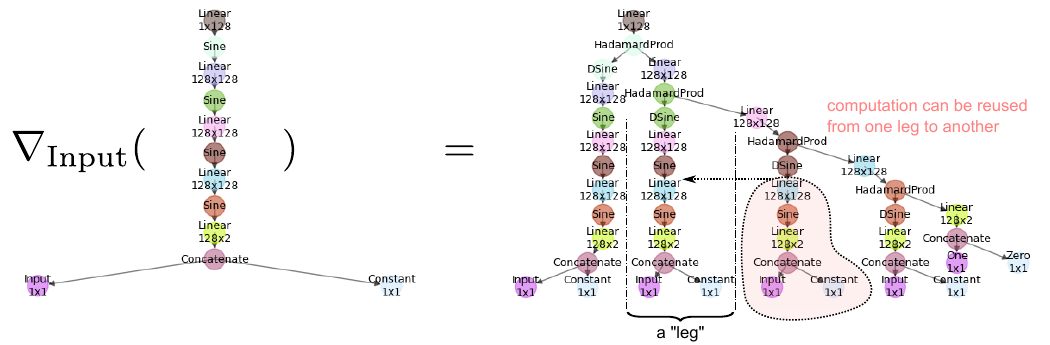}
    \caption{Visualizations of integral and grad networks generated with the AutoInt graph implementation. Directed acyclic graphs corresponding to the integral network (left) and grad network (right) consist of computational nodes and their dependencies. Here, arrows point from a node to each of its dependencies, indicating which node should be computed first. The grad network has a tree-like structure, and computes the partial derivative of the integral network with respect to one of its inputs. Nodes within the ``legs'' of the grad network appear multiple times, thus these computations can be re-used during a forward pass in order to improve performance.}
	\label{fig:supp_computation_graphs}
    \vspace{-1em}
\end{figure*}

\subsection{Overview}
In the AutoInt framework, we start by specifying the architecture of the integral network: the number of layers, features, the type of non-linearities, and the input parameterization.
Our implementation of AutoInt relies on a evaluating computational graphs, where dependencies are modeled using directed acyclic graphs (DAGs).  
With this graph-based representation, we create an automated pipeline for instantiating grad networks from integral networks, and we develop an efficient procedure for evaluating grad networks during training.
\vspace{-1em}
\paragraph{DAGs to represent computational graphs.} Our AutoInt implementation internally maintains the computational graph of neural networks as a Directed Acyclic Graphs (DAG).
Most nodes in the graph represent computational operators and there are two kinds of leaf nodes: (1) an input node with respect to which we can take the derivative of the graph and (2) a constant input node.
Directed edges represent dependencies between nodes and point towards dependencies (i.e., other nodes to be computed first). 
Hence, nodes of in-degree zero are final results of the computation graph.
\vspace{-1em}
\paragraph{Building the grad network.} To instantiate a \emph{grad network}, AutoInt performs auto differentiation on the DAG of the integral network. Each node is called in a topological order and provides its own derivative. This recursive chain of calls builds the computation graph of the grad network. 
\vspace{-1em}
\paragraph{Evaluating the grad network.} Once the grad network is built it can be evaluated using a reverse topological ordering of its nodes: starting from the leaves and tracing computation back to the root(s). 
Note that this procedure is different than backpropagation, where intermediate results from the forward pass are stored in order to evaluate the graph associated with the backward pass.
In AutoInt, given the grad network is a separate entity from the integral network, the intermediate results from the forward pass are unavailable. 
However, the grad network can still be computed efficiently by reusing computations from the ``legs'' of the network, since these nodes share weights and perform the same computations (see Figure~\ref{fig:supp_computation_graphs}). 
This is done by maintaining a lexicographic ordering between nodes of same in-degree in the topological ordering.
The lexicographic ordering is defined by the order in which nodes were created during differentiation of the integral network: nodes created last appear first in the ordering.
Thus the evaluation of the grad network in forward mode proceeds by calling each node in this lexicographic-topological ordering. Nodes save their computation in a cumulative fashion; as the legs of the network are computed, the last results are kept to be reused in other legs.
 
\paragraph{Training the grad network.} The weights of the grad network are trained with backpropagation. 
During the evaluation of the grad network, computations are also saved for the backward pass performed by the backpropagation algorithm. 
The weights of the grad network are then updated with Adam~\cite{kingma2014adam}, a variant of SGD.
\begin{figure*}[h]
	\includegraphics[width=\textwidth]{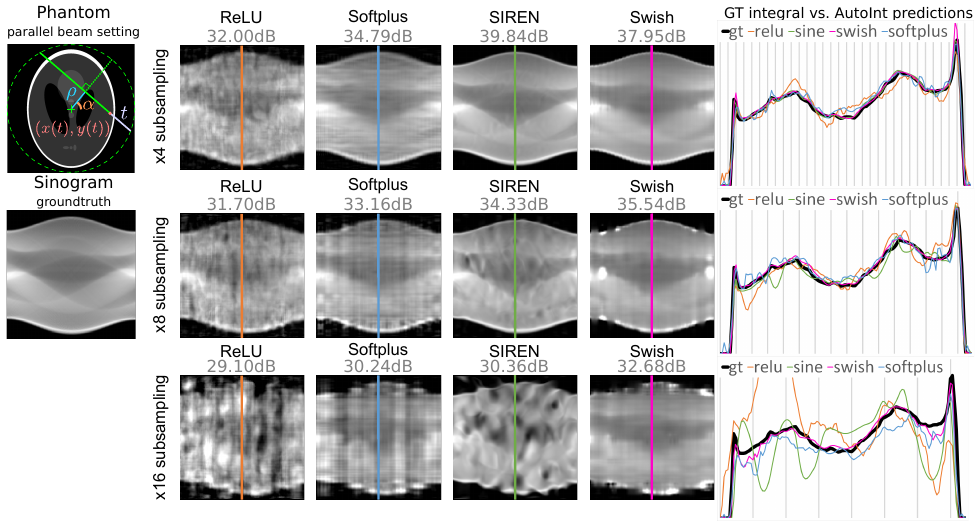}
    \caption{Supplemental results of AutoInt for computed tomography. Left: illustration of the parameterization. Center: sinograms computed with the integral networks using different nonlinear activation functions. The ground truth (GT) sinogram is subsampled in angle by 4$\times$ (top), 8$\times$ (middle), and 16$\times$ (bottom). The optimized networks are used to interpolate the missing measurements. Using the Swish activation performs best in these experiments. Right: 1D scanlines of the sinogram centers shows the interpolation behavior of each method for each subsampling level.}
	\label{fig:supp_tomography}
    \vspace{-1em}
\end{figure*}
\subsection{Implementation and Results}
We implemented AutoInt in Python and PyTorch~\cite{paszke2019pytorch}, and we used the Python Networkx library to maintain the data structures forming the backend of our computational graphs. 
At the core of our AutoInt implementation is a custom AutoDiff tool that acts on computational nodes defined by the framework. 
After an integral network is built using the computational nodes, it is parsed by our AutoDiff tool, and the grad network is created. 
Importantly, each of our computational nodes wraps a Pytorch module, enabling parameter sharing within and between the networks. 
After the networks are instantiated, it is convenient to use the Pytorch AutoDiff to perform backpropagation and weight updates during training. 
Reassembling the integral network is trivial due to the weight sharing mechanism; the integral and grad network parameters always match as they are updated during training. 

AutoInt can also be directly implemented in Pytorch, though with relatively severe performance penalties.
To train the grad network, one would perform the following steps during each training iteration: (1) compute the output of the integral network, (2) use AutoDiff to calculate the derivative of the output with respect to the input variable of integration, and (3) perform backpropagation on a loss calculated using the derivative. 
This procedure is inefficient because, compared to our framework, it requires an extra forward pass through the integral network, and it requires re-assembling the grad network at every training iteration.  
Our AutoInt implementation compares favorably to this direct PyTorch implementation.
We measure a savings of over $15\%$ in GPU memory, and a more than $1.8$x speedup in the number of training iterations per second for the volume rendering task described in the main paper.

\section{Supplemental Results}
\subsection{Sparse-View Computed Tomography}

We include supplemental results for the sparse-view computed tomography task described in the main paper.
Here, we train a grad network on sparse angular projections (i.e., a subsampled sinogram) of a 2-dimensional phantom.
After training, the integral network is evaluated for all projection angles to inpaint the missing regions of the sinogram.
As shown in Figure~\ref{fig:supp_tomography}, we find that using the Swish non-linearity in the integral network results in the best generalization performance in term of inpainting the unseen projections. 
SIREN~\cite{sitzmann2020siren}, which uses the sine nonlinearity, performs well in the densely supervised case, but performs increasingly degrades for inpainting the sinograms for 8$\times$ and 16$\times$ angular subsampling. 
We also show the performance of ReLU and Softplus.
Both of these non-linearities produce relatively noisy inpainted results. 
ReLU is particularly unsuited to learning this representation, as its derivative, which appears in the grad network, has a zero-valued derivative almost everywhere.  

\begin{figure}[h!]
	\includegraphics[]{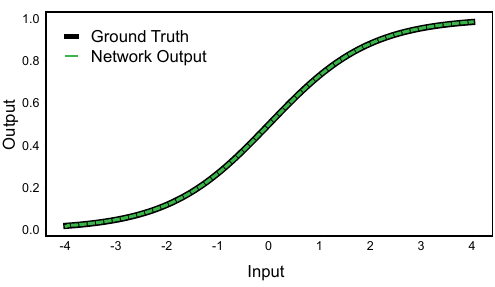}
    \caption{\highlight{Supervising the integral network with definite integrals. We set an antiderivative $F$ to be equal to a sigmoid function and supervise an integral network with values of definite integrals as described by Equation~\ref{eqn:definite}. The network recovers the antiderivative $F$ up to a scalar constant and is compared to ground truth (we remove the offset for visualization).}}
    \label{fig:definite}
    \vspace{-1em}
\end{figure}

\highlight{
\subsection{Supervising the Integral Network}
In the main paper, we use AutoInt to supervise the grad network and then reconstruct the integral network. It is also possible to directly supervise the integral network, $\imp_\param$, to learn an antiderivative. One way to do this is with a provided dataset of definite integrals. For example, consider we wish to learn an antiderivative $F(x)$ over an interval $[a, b]$. In this case, we can supervise the integral network directly over definite integrals $F(x_2) - F(x_1) = \int_{x_1}^{x_2} f(x) \mathrm{d}x$ for $x_1, x_2 \in [a, b]$. Here, the integral network is trained to minimize a loss function of the following form.
\begin{equation}
    \text{arg min}_\theta\: \lVert \left[\imp_\param(x_2) - \imp_\param(x_1)\right] - \left[ F(x_2) - F(x_1)\right] \rVert_2^2.
    \label{eqn:definite}
\end{equation}
}
\highlight{We demonstrate training a neural network with Swish non-linearities to learn a sigmoid function $F(x) = 1/(1 + e^{-x})$ from a dataset of randomly sampled definite integrals as described by Equation~\ref{eqn:definite}. The resulting fit is shown in Figure~\ref{fig:definite}. The network accurately fits the sigmoid function up to a scalar constant (which we remove in the visualization of Figure~\ref{fig:definite}).} 

\subsection{Results on Captured Data}
\begin{figure*}[h!]
	\includegraphics[width=\textwidth]{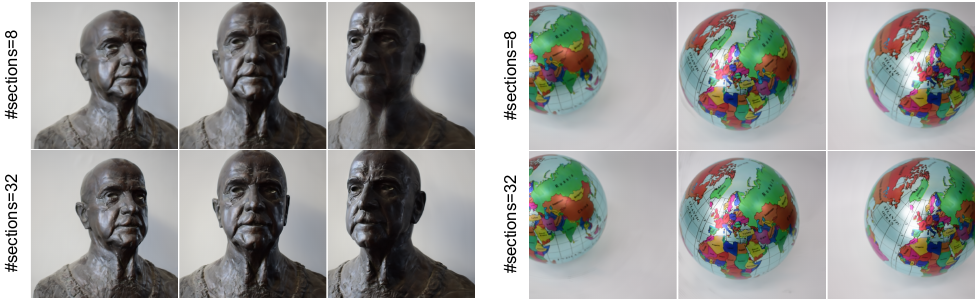}
    \caption{Supplemental results of AutoInt for real captures. We show qualitative results of AutoInt with 8 and 32 sections on the \textit{Statue} and the \textit{Globe} scenes used in DeepVoxels \cite{sitzmann2019deepvoxels}.}
    \label{fig:deepvoxels}
\end{figure*}

In Figure~\ref{fig:deepvoxels}, we include supplemental results using real captured data from DeepVoxels~\cite{sitzmann2019deepvoxels}.
The scenes were trained on half of the randomly sampled images of the provided RGB data and are shown here for a held out test set.
Both AutoInt results using 8 and 32 sections achieve similarly high image quality on these captured scenes.

\begin{figure*}[h!]
    \includegraphics[width=\textwidth]{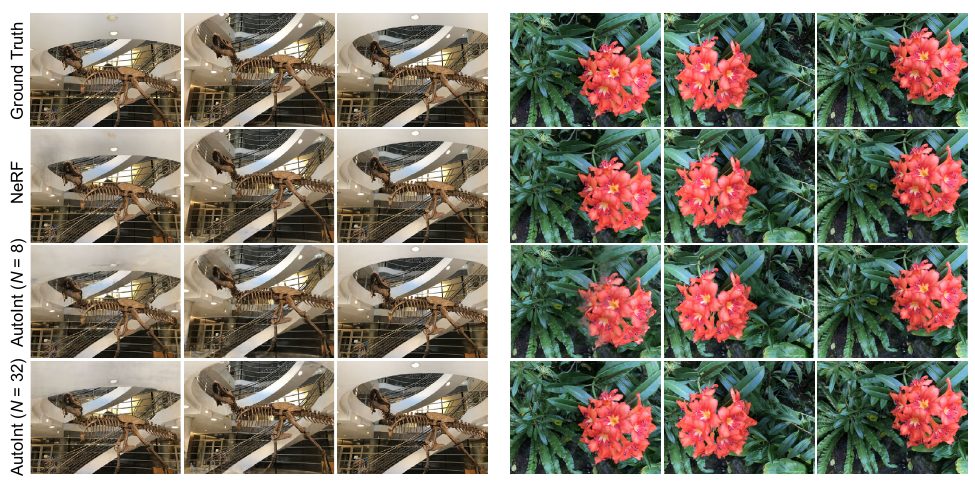}
    \caption{\highlight{Supplemental results of AutoInt for real captures from the Local Light Field Fusion datasets~\cite{Mildenhall:2019}. We show qualitative results of AutoInt with 8 and 32 sections on the \textit{T-Rex} and \textit{Flower} scenes.}}
    \label{fig:llff}
\end{figure*}

\highlight{We also show results from the Local Light Field Fusion (LLFF) datasets~\cite{Mildenhall:2019} and qualitative comparisons between NeRF~\cite{mildenhall2020nerf} and AutoInt in Figure~\ref{fig:llff}.
The LLFF datasets consist of captured, forward-facing image data, and we train and evaluate all images at a resolution of 378 by 504 pixels.
We train with a batch size of one and otherwise use the same training parameters as the Blender datasets as described in the main text.
Following the LLFF authors, we partition the captured images so that 1/8 of the images are used for the test set.
In Table~\ref{tab:supp_llff}, we show quantitative comparisons between AutoInt and NeRF for the LLFF datasets. }

\subsection{Synthetic Blender Dataset Scenes}
\highlight{In Table~\ref{tab:supp_blender} we provide additional quantitative evaluations using peak signal-to-noise ratio (PSNR), the structural similarity index measure (SSIM)~\cite{wang2004image}, and the learned perceptual image patch similarity (LPIPS) metric~\cite{zhang2018unreasonable}.} 

\begin{table*}
    \centering
	\begin{tabular}{lcccccccc}
		\toprule[0.8pt]
        & \multicolumn{8}{c}{PSNR$\uparrow$}\\ 
        & Room & Fern & Leaves & Fortress & Orchids & Flower & T-Rex & Horns \\
        \midrule[0.4pt]
        NeRF~\cite{mildenhall2020nerf} &  \textbf{33.60} & \textbf{26.92} & \textbf{22.50} & \textbf{32.94} & \textbf{21.37} & \textbf{28.57} & \textbf{28.26} & \textbf{29.26} \\
        AutoInt ($N$=8)                & 28.33 & 22.11 & 19.61 & 28.63 & 16.85 & 26.65 & 24.90 & 26.01 \\
        AutoInt ($N$=16)               & 29.97 & 23.29 & 18.78 & \underline{29.53} & \underline{17.71} & 27.60 & 25.58 & 26.72 \\
        AutoInt ($N$=32)               & \underline{30.72} & \underline{23.51} & \underline{20.84} & 28.95 & 17.30 & \underline{28.11} & \underline{27.18} & \underline{27.64} \\
		\bottomrule[0.8pt]
	\end{tabular}
    \vspace{2em}

    \centering
	\begin{tabular}{lcccccccc}
		\toprule[0.8pt]
        & \multicolumn{8}{c}{SSIM$\uparrow$}\\ 
        & Room & Fern & Leaves & Fortress & Orchids & Flower & T-Rex & Horns \\
        \midrule[0.4pt]
        NeRF~\cite{mildenhall2020nerf} & \textbf{0.980} & \textbf{0.903} & \textbf{0.851} & \textbf{0.962} & \textbf{0.800} & \textbf{0.931} & \textbf{0.953} & \textbf{0.947} \\
        AutoInt ($N$=8)                & 0.941 & 0.771 & 0.745 & 0.896 & 0.560 & 0.882 & 0.888 & 0.880 \\
        AutoInt ($N$=16)               & 0.954 & 0.802 & 0.712 & \underline{0.914} & \underline{0.607} & 0.903 & 0.904 & 0.894 \\
        AutoInt ($N$=32)               & \underline{0.966} & \underline{0.810} & \underline{0.795} & 0.910 & 0.583 & \underline{0.917} & \underline{0.931} & \underline{0.908} \\
		\bottomrule[0.8pt]
	\end{tabular}
    \vspace{2em}

    \centering
	\begin{tabular}{lcccccccc}
		\toprule[0.8pt]
        & \multicolumn{8}{c}{LPIPS$\downarrow$}\\ 
        & Room & Fern & Leaves & Fortress & Orchids & Flower & T-Rex & Horns \\
        \midrule[0.4pt]
        NeRF~\cite{mildenhall2020nerf} & \textbf{0.038} & \textbf{0.085} & \textbf{0.103} & \textbf{0.024} & \textbf{0.108} & \textbf{0.057} & \textbf{0.049} & \textbf{0.058} \\
        AutoInt ($N$=8)                & 0.110 & \underline{0.276} & 0.171 & 0.092 & 0.313 & 0.113 & 0.123 & 0.213 \\
        AutoInt ($N$=16)               & 0.102 & 0.283 & 0.218 & \underline{0.086} & \underline{0.268} & 0.090 & 0.107 & 0.202 \\
        AutoInt ($N$=32)               & \underline{0.075} & 0.277 & \underline{0.156} & 0.107 & 0.302 & \underline{0.075} & \underline{0.080} & \underline{0.177} \\
		\bottomrule[0.8pt]
	\end{tabular}
    \vspace{1em}
    \caption{\highlight{Per-scene quantitative results calculated across the test sets of the Local Light Field Fusion datasets~\cite{Mildenhall:2019}.}}
	\label{tab:supp_llff}
\end{table*}

\begin{table*}
    \centering
	\begin{tabular}{lcccccccc}
		\toprule[0.8pt]
        & \multicolumn{8}{c}{PSNR$\uparrow$}\\ 
        & Chair & Drums & Ficus & Hotdog & Lego & Materials & Mic & Ship \\
        \midrule[0.4pt]
        NeRF~\cite{mildenhall2020nerf} & \textbf{33.00} & \textbf{25.01} & \textbf{30.13} & \textbf{36.18} & \textbf{32.54} & \textbf{29.62} & \textbf{32.91} & \textbf{28.65} \\
        NV~\cite{Lombardi:2019} & \underline{28.33} & \underline{22.58} & 24.79 & 30.71 & 26.08 & 24.22 & 27.78 & 23.93 \\
        AutoInt ($N$=8) & 25.60 & 20.78 & 22.47 & \underline{32.33} & 25.09 & 25.90 & 28.10 & 24.15 \\
        AutoInt ($N$=16) & 25.65 & 21.30 & 23.95 & 31.28 & 25.48 & 28.05 & 28.36 & 24.26 \\
        AutoInt ($N$=32) & 25.82 & 22.02 & \underline{25.51} & 31.84 & \underline{27.26} & \underline{28.58} & \underline{28.42} & \underline{25.18}\\
		\bottomrule[0.8pt]
	\end{tabular}
	
    \vspace{2em}

    \centering
	\begin{tabular}{lcccccccc}
		\toprule[0.8pt]
        & \multicolumn{8}{c}{SSIM$\uparrow$}\\ 
        & Chair & Drums & Ficus & Hotdog & Lego & Materials & Mic & Ship \\
        \midrule[0.4pt]
        NeRF~\cite{mildenhall2020nerf} & \textbf{0.967} & \textbf{0.925} & \textbf{0.964} & \textbf{0.974} & \textbf{0.961} & \textbf{0.949} & \textbf{0.980} & \underline{0.856} \\
        NV~\cite{Lombardi:2019} & 0.916 & 0.879 & 0.910 & 0.944 & 0.880 & 0.888 & 0.946 & 0.784 \\
        AutoInt ($N$=8) & \underline{0.928} & 0.861 & 0.898 & \textbf{0.974} & 0.900 & 0.930 & 0.948 & 0.852 \\
        AutoInt ($N$=16) & 0.925 & 0.869 & 0.909 & 0.971 & 0.905 & 0.947 & \underline{0.951} & 0.853 \\ 
        AutoInt ($N$=32) & 0.926 & \underline{0.885} & \underline{0.926} & \underline{0.973} & \underline{0.929} & \underline{0.953} & \underline{0.951} & \textbf{0.869} \\
		\bottomrule[0.8pt]
	\end{tabular}
    \vspace{2em}

    \centering
	\begin{tabular}{lcccccccc}
		\toprule[0.8pt]
        & \multicolumn{8}{c}{LPIPS$\downarrow$}\\ 
        & Chair & Drums & Ficus & Hotdog & Lego & Materials & Mic & Ship \\
        \midrule[0.4pt]
        NeRF~\cite{mildenhall2020nerf} & \textbf{0.046} & \textbf{0.091} & \textbf{0.044} & 0.121 & \textbf{0.050} & \textbf{0.063} & \textbf{0.028} & \textbf{0.206} \\
        NV~\cite{Lombardi:2019} & \underline{0.109} & 0.214 & 0.162 & 0.109 & 0.175 & 0.130 & \underline{0.107} & \underline{0.276} \\
        AutoInt ($N$=8) & 0.141 & 0.224 & 0.148 & \textbf{0.080} & 0.175 & 0.136 & 0.131 & 0.323 \\
        AutoInt ($N$=16) & 0.149 & 0.221 & 0.139 & 0.095 & 0.171 & 0.110 & 0.130 & 0.320 \\
        AutoInt ($N$=32) & 0.149 & \underline{0.209} & \underline{0.109} & \underline{0.088} & \underline{0.135} & \underline{0.100} & 0.127 & 0.295 \\
		\bottomrule[0.8pt]
	\end{tabular}
    \vspace{1em}
	\caption{Per-scene quantitative results calculated across the test sets of the synthetic Blender datasets. These simulated scenes contain challenging geometries and reflectance properties.}
	\label{tab:supp_blender}
\end{table*}

{\small
\bibliographystyle{ieee_fullname}
\bibliography{references}
}